\documentclass[sigconf,pbalance,authorversion]{acmart-preprint}
\AtBeginDocument{%
  }

\copyrightyear{2026}
\acmYear{2026}
\setcopyright{cc}
\setcctype{by}
\acmConference[ICCAD '26]{IEEE/ACM International Conference on Computer-Aided Design}{November 08--12, 2026}{San Jose, CA, USA}
\acmBooktitle{IEEE/ACM International Conference on Computer-Aided Design (ICCAD '26), November 08--12, 2026, San Jose, CA, USA}
\acmDOI{10.1145/3831252.3834266}
\acmISBN{979-8-4007-2873-0/2026/11}

\newtheorem{task}{Task}
\newtheorem{definition}{Definition}

\usepackage{algorithm}
\usepackage{algorithmic}
\usepackage{dsfont}
\usepackage{tikz}
\usepackage{subcaption}
\usepackage{multirow}

\usepackage[para,online,flushleft]{threeparttable}

\begin{document}

\title{GNN-NPN: Efficient NPN Classification for Large Functions using Graph Neural Network}
\settopmatter{authorsperrow=4}

\author{Chengdi Cao}
\email{chengdicao@ucla.edu}
\affiliation{%
  \institution{University of California, Los Angeles}
  \city{Los Angeles}
  \state{CA}
  \country{USA}}

\author{Hanyu Wang}
\email{hanyuwang@ucla.edu}
\affiliation{%
  \institution{University of California, Los Angeles}
  \city{Los Angeles}
  \state{CA}
  \country{USA}}

\author{Wan-Hsuan Lin}
\email{wanhsuanlin@ucla.edu}
\affiliation{%
  \institution{University of California, Los Angeles}
  \city{Los Angeles}
  \state{CA}
  \country{USA}}

\author{Jason Cong}
\email{cong@cs.ucla.edu}
\affiliation{%
  \institution{University of California, Los Angeles}
  \city{Los Angeles}
  \state{CA}
  \country{USA}}

\renewcommand{\shortauthors}{Cao et al.}

\begin{abstract}

Two Boolean functions are considered NPN equivalent if one can be transformed into the other by negating inputs, permuting inputs, or negating the output.
Efficient NPN classification is valuable in fields such as logic synthesis, quantum computing, and information theory.
However, exact NPN classification demands exhaustive enumeration of all possible input permutations and negations. Its complexity grows exponentially, making it impractical for larger functions with more inputs.
In this paper, we introduce the first learning-based formulation for NPN classification by recasting it as NPN action prediction.
Our method represents Boolean functions as AND-Inverter Graphs (AIGs) and applies a bidirectional heterogeneous graph neural network to predict input negations, output negation, and input permutations.
To mitigate the non-canonicity of logic networks, we further introduce graph rewriting techniques that reduce structural ambiguity and recover useful symmetries before prediction.
Experimental results show that NPN action prediction is learnable, that the resulting canonicalization quality improves over heuristic baselines, and that on practical benchmarks our method approaches exact methods on tractable cases while scaling to functions with up to 16 inputs.

\end{abstract}



\maketitle

\section{Introduction}

Two Boolean functions are \emph{NPN equivalent} if one can be transformed into the other through a combination of three operations: input negation, input permutation, and output negation~\cite{goto1962some, muroga1979logic}.
NPN classification is the task of classifying Boolean functions into NPN-equivalent classes, and is a fundamental problem in logic synthesis.
It is also of practical significance in a wide range of applications, including logic optimization~\cite{mishchenko2006dag, rudell1993dynamic, yang2012lazy}, technology mapping~\cite{chatterjee2006reducing, huang2013fast, cong1994flowmap, kennings2010efficient}, and emerging technologies such as circuit synthesis for quantum computing~\cite{wang2024quantum}.

However, NPN classification is computationally challenging due to the combinatorial explosion of possible transformations as the number of input bits increases.
Specifically, for an \(n\)-bit Boolean function, there are \(2^n\) input negation patterns, \(n!\) input permutations, and \(2\) output negations, yielding a total of \(2^{n+1} \cdot n!\) possible NPN transformations.
This exponential growth makes exact NPN classification intractable for functions with more than a few inputs.

Depending on the approach, NPN classifiers can be broadly categorized as either \emph{exact} or \emph{heuristic}.
Exact methods return provably correct canonical representatives but are typically limited to functions with only a few inputs~\cite{rudell1993dynamic, li2011aig} due to their computational demands.
To enable classification of larger functions, a number of heuristic approaches have been proposed.
These methods trade optimality for scalability: they attempt to reduce the search space and often work well in practice, but they do not provide the same guarantees as exact methods~\cite{huang2013fast,soeken2016heuristic, petkovska2016fast, zhou2019fast}.

This exact-versus-heuristic distinction does not fully capture the practical scalability bottleneck, because the underlying representation also matters.
Many existing methods operate on truth tables or on features derived from truth tables.
Such methods can prune the NPN search space using function symmetries~\cite{chai2006building, abdollahi2008symmetry, agosta2009transform, zhou2019fast}, functionality-based signatures~\cite{zhang2023rethinking}, or local search~\cite{huang2013fast}, but they still rely on the truth table representation whose size grows exponentially with the number of inputs.
As a result, even heuristic methods remain increasingly expensive on larger functions.

In contrast, logic networks such as AND-Inverter Graphs (AIGs) provide a more scalable representation for practical synthesis flows.
AIGs are compact, standard in modern synthesis tools, and expose structural regularities such as symmetry that are relevant to NPN transformations.
{But} AIGs are non-canonical: the same Boolean function can be implemented by different graphs.
This representation ambiguity poses challenges for learning on logic networks.

In this paper, we propose GNN-NPN, a novel heuristic approach to NPN classification that leverages both data-driven machine learning and structural cues from graph representations of Boolean functions.
Specifically, we formulate NPN classification as a problem of predicting \emph{NPN actions} that move a Boolean function toward its canonical representative.
We represent Boolean functions as AIGs and design a graph neural network (GNN) to learn structural and functional features from these graphs for action prediction.
To mitigate the challenges posed by the non-canonical AIG representation, we further introduce graph rewriting techniques that reduce structural ambiguity and recover useful symmetries before feeding the graphs into the GNN.
Together, these components yield a scalable and high-performing heuristic-based workflow for NPN classification on large Boolean functions.

The main contributions of this paper are listed as follows:
\begin{enumerate}
    \item We formulate NPN classification as a machine learning problem by casting it as NPN action prediction. This formulation clarifies the relationship among classification, canonicalization, and action prediction, while ensuring that every model output remains a valid NPN transformation (Section~\ref{sec:problem-formulation}).
    \item We propose GNN-NPN, an AIG-based action-predictor built on a bidirectional heterogeneous GNN. The model predicts input negations, output negation, and input permutations directly from logic-network structure, enabling graph-based NPN canonicalization on larger functions (Section~\ref{sec:gnn-classifier}).
    \item We develop a graph rewriting flow for non-canonical AIGs that recovers useful symmetries and reduces structural ambiguity before prediction, improving the consistency of graph representations used by the GNN (Section~\ref{sec:graph-representation}).
    \item We perform extensive experiments to evaluate the method at the levels of action prediction, canonicalization, and practical classification. The results show that GNN-NPN improves canonicalization quality over heuristic baselines, approaches exact methods on tractable problem sizes, and scales to functions with up to 16 inputs (Section~\ref{sec:exp}).
\end{enumerate}

{The source code and scripts are available at \href{https://github.com/UCLA-VAST/GNN-NPN}{{\nolinkurl{https://github.com/UCLA-VAST/GNN-NPN}}}.}

\section{Background}
\label{sec:background}

\subsection{Boolean Function and Its Representations}

A Boolean function with \(n\) inputs is a mapping
\begin{equation}
    f:\mathbb{B}^n \rightarrow \mathbb{B},
\end{equation}
where \(\mathbb{B}=\{0,1\}\).
In logic networks, the \(n\) input bits correspond to primary inputs (PIs) and the output bit corresponds to the primary output (PO).
{We denote the input vector by \(\Vec{x}=(x_0,x_1,\dots,x_{n-1})\in\mathbb{B}^n\).}
The input variables are ordered from the most significant bit (MSB) \(x_0\) to the least significant bit (LSB) \(x_{n-1}\).

A \emph{truth table} is a canonical representation of a Boolean function, since there exists a bijective correspondence between Boolean functions and their truth tables.
Truth tables can be denoted as binary strings and the corresponding values are called the lexicographic values of Boolean functions, denoted \(\mathrm{lex}(f)\). 
For example, 
the truth table of $f_1 = (x_0 \vee x_1) \wedge x_2$
is \texttt{10101000} and \(\mathrm{lex}(f_1)=168\).
Truth tables are convenient for defining canonical representatives and for implementing exact algorithms on small functions.
However, their size grows exponentially with the number of inputs, making them increasingly expensive for large Boolean functions.

In practice, \emph{logic networks}, modelled as directed acyclic graphs (DAGs), are more compact for representing large Boolean functions and are more commonly used in logic synthesis workflows.
Specifically, \emph{AND-Inverter Graphs} (AIGs), which consist of two-input \texttt{AND} gates and inverters, have become a de facto standard representation in modern synthesis tools~\cite{brayton2010abc}.
However, they are non-canonical: the same Boolean function can be implemented by different logic networks.
This tradeoff between scalability and non-canonicity motivates the graph-based learning approach in this paper.

\subsection{NPN Equivalence and Canonical Representative}

For brevity, throughout this paper, we denote an NPN transformation {\(a\)} consisting of input negation \(p\in\mathbb{B}^n\), input permutation \(\pi\in\Pi_n\) and output negation \(q\in\mathbb{B}\) as \(a=(p,\pi,q)\), where \(\Pi_n\) is the set of \(n\)-element permutations.
For an input vector \(\Vec{x}=(x_0,x_1,\dots,x_{n-1})\), the permutation \(\pi\) acts as
\begin{equation}
    \pi(\Vec{x}) = \left(x_{\pi(0)}, x_{\pi(1)}, \dots, x_{\pi(n-1)}\right).
\end{equation}
Applying \(a\) to a Boolean function \(f\) yields
\begin{equation}
    f'=a(f)
    \quad\mathrm{s.t.}\quad
    f'(\Vec{x}) = q \oplus f(\pi(p \oplus \Vec{x})),
\end{equation}
where \(\oplus\) is the bitwise XOR operation.

We define the NPN equivalence relationship as follows:

\begin{definition}[NPN Equivalence]
    Two Boolean functions $f_1$ and $f_2$ are NPN-equivalent if and only if one can be transformed into the other through an NPN transformation,
    \begin{equation}
        f_1 \sim_{\text{NPN}} f_2 \quad\Longleftrightarrow\quad \exists a\quad\mathrm{s.t.}\quad f_2=a(f_1).
    \end{equation}
\end{definition}

Examples of NPN-equivalent Boolean functions can be found in Figure~\ref{fig:gnn-npn-workflow}, where {\(f_1 = \texttt{0xa8}\) can be transformed into \(f_1' = \texttt{0x07}\)} by applying \(x_0' = \neg x_2\), \(x_1' = \neg x_0\), and \(x_2' = \neg x_1\).

Given logic network implementations of a Boolean function, all Boolean functions that are NPN-equivalent to it can be implemented from the same logic networks by permuting the input bits and adding/removing inverters.
Therefore, in logic synthesis workflows, it is sufficient to optimize only one canonical representative for each NPN-equivalent class, which can drastically reduce the size of precomputed libraries.
These distinguished Boolean functions are called the canonical representatives of their NPN-equivalent classes.
In this paper, we define them in terms of lexicographic value:

\begin{definition}[Canonical Representative]
    \label{def:canonical-representative}
    {The canonical representative of the NPN-equivalence class containing
    \(f\) is the unique Boolean function \(\hat{f}\) satisfying}
    \begin{equation}
        {
            \hat{f} \sim_{\mathrm{NPN}} f
            \quad\text{and}\quad
            \mathrm{lex}(\hat{f})
            =
            \min_{f' \sim_{\mathrm{NPN}} f} \mathrm{lex}(f')
        }.
    \end{equation}
\end{definition}

\subsection{Graph Neural Network}

Since logic networks are naturally represented as graphs, graph neural networks (GNNs) are a natural model family for this domain.
GNNs operate on graph-structured data by iteratively propagating and aggregating information over nodes and edges~\cite{gori2005new}.

DAG-structured GNNs specialize this idea to directed acyclic graphs by following the topological order of the graph, which matches the computational structure of AIGs.
Such models have already shown promise in logic synthesis tasks including logic optimization~\cite{zhu2020exploring}, symbolic reasoning on Boolean networks~\cite{wu2023gamora}, and functionality-aware circuit representation learning~\cite{shi2023deepgate2}.

Existing heuristic methods often rely on explicitly engineered cues such as symmetry information~\cite{chai2006building, abdollahi2008symmetry, agosta2009transform, zhou2019fast} and function signatures~\cite{zhang2023rethinking} to guide canonicalization.
A GNN offers a complementary learned mechanism: it can exploit symmetries already present in the graph and can also learn distributed representations that play a role similar to structural or functional signatures, without restricting the method to a fixed set of handcrafted features.

\section{Formulating NPN Classification as a Machine Learning Problem}
\label{sec:problem-formulation}
\begin{figure*}[tb]
    \centering
    \includegraphics[width=0.9\textwidth]{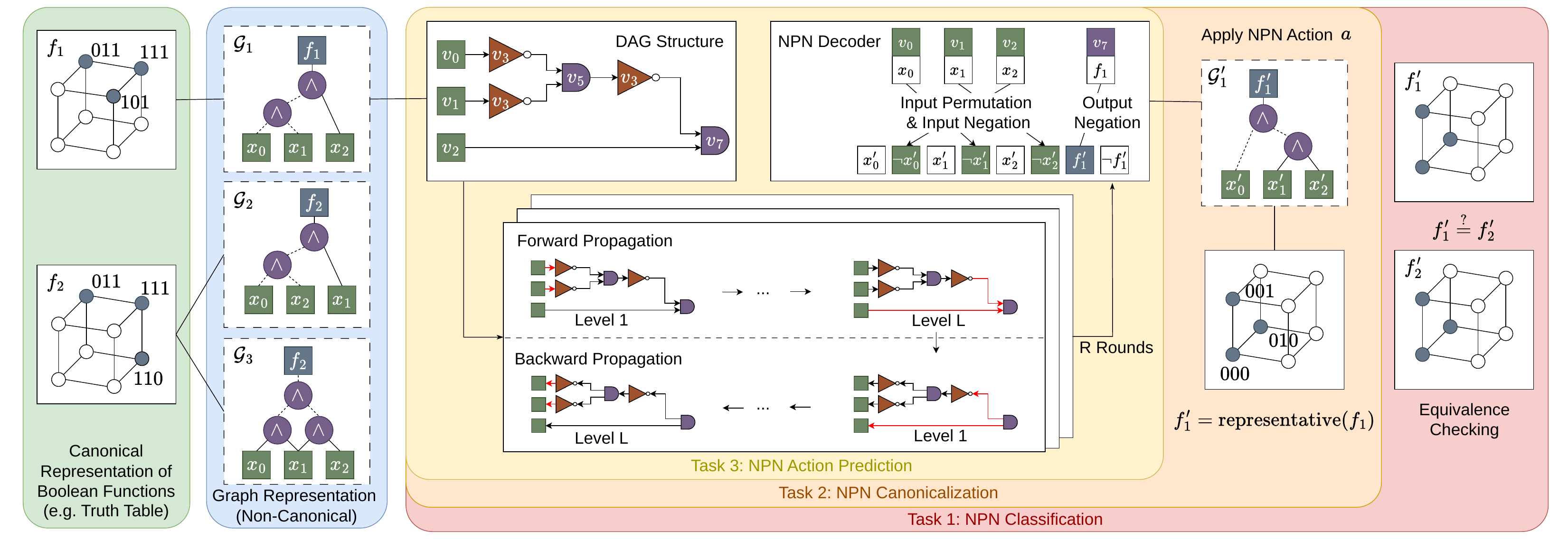}
    \caption{Overview of the proposed GNN-NPN Workflow. At its core, our approach employs an NPN action predictor (introduced in Section~\ref{sec:gnn-classifier}) that takes AIG as input and produces an NPN action. We demonstrate in Section~\ref{sec:problem-formulation} that this model can be utilized to {perform NPN canonicalization for a given function} and consequently, perform NPN classification. Additionally, we propose graph rewriting algorithms (detailed in Section~\ref{sec:graph-representation}) to enhance the symmetry in AIG representations.}
    \label{fig:gnn-npn-workflow}
\end{figure*}

For a fixed input size \(n\), let \(\mathcal{F}_n\) denote the set of all \(n\)-input Boolean functions, and let
\begin{equation}
    \label{eq:action-space}
    \mathcal{A}_n = \mathbb{B}^n \times \Pi_n \times \mathbb{B}
\end{equation}
denote the set of all \(n\)-input NPN transformations.
The goal of this paper is to perform NPN classification, which we denote as Task~\ref{task:cls}.

\begin{task}[NPN classification]
    \label{task:cls}
    Given two Boolean functions \(f_1,f_2\in\mathcal{F}_n\), determine whether they are NPN-equivalent,
    \begin{equation}
        h:\mathcal{F}_n\times\mathcal{F}_n\rightarrow \{0,1\},
        \quad
        h(f_1,f_2)=
        \begin{cases}
            1 & f_1\sim_{\text{NPN}}f_2 \\[4pt]
            0 & \text{otherwise}
        \end{cases}.
    \end{equation}
\end{task}

{Most existing methods perform NPN canonicalization by finding the canonical representative}~\cite{huang2013fast,petkovska2016fast,zhou2019fast}, which we denote as Task~\ref{task:canon}.

\begin{task}[{NPN Canonicalization}]
    \label{task:canon}
    Given a Boolean function \(f\in\mathcal{F}_n\), find its canonical representative \(\hat{f}\) {as defined in Definition~\ref{def:canonical-representative}},
    \begin{equation}
        r^\star:\mathcal{F}_n\rightarrow\mathcal{F}_n,
        \quad
        r^\star(f)=\hat{f}.
    \end{equation}
\end{task}

When \(r^\star\) is available, Task~\ref{task:cls} reduces to equality verification:

\begin{equation}
    h(f_1, f_2) = \mathds{1}\left\{r^\star(f_1) = r^\star(f_2)\right\}.
\end{equation}

Although Tasks~\ref{task:cls} and~\ref{task:canon} have well-defined inputs and outputs, they are not the most suitable interfaces for machine learning.
A learned \(\tilde{h}\) may mistaken non-equivalent pairs as equivalent.
Similarly, a learned \(\tilde{r^\star}\) may output a function that is not even NPN-equivalent to the input.
In contrast, if the learning module is restricted to predicting elements of \(\mathcal{A}_n\) {as defined in Equation~\ref{eq:action-space}}, then every model output is a valid NPN transformation and therefore preserves NPN equivalence by construction.
This motivates the following action-prediction formulation.

\begin{task}[NPN Action Prediction]
    \label{task:action}
    Given a Boolean function \(f\in\mathcal{F}_n\), predict an NPN transformation \(a\in\mathcal{A}_n\) that maps \(f\) toward its canonical representative.
    Ideally, the predicted action solves
    \begin{equation}
        a^\star(f)\in\arg\min_{a\in\mathcal{A}_n}\mathrm{lex}\left(a(f)\right),
    \end{equation}
    so that {the} predicted action maps \(f\) to the canonical representative \(\hat{f}\).
\end{task}

The relationship between the three tasks is depicted in Figure~\ref{fig:gnn-npn-workflow}.
To the best of our knowledge, this is the first machine learning formulation for NPN classification.
In practice, our learned model approximates the ideal action predictor in Task~\ref{task:action}.
The predicted action need not be optimal in one step; it only needs to remain within the correct NPN class and move the function toward a better representative, after which local refinement can be applied.
We highlight two advantages of this formulation:
\begin{enumerate}
    \item The transformed function remains correct by construction. Because every predicted output is a valid NPN transformation, applying it to \(f\) always produces a function that remains in the same NPN-equivalence class as \(f\).
    \item The task tolerates imperfect predictions. The learned predictor does not need to recover the exact minimizer \(a^\star(f)\) in one step; suboptimal but valid actions can still reduce the lexicographic value and can be {further} refined by post{-}processing such as local search.
\end{enumerate}

\section{GNN-NPN Action Predictor}
\label{sec:gnn-classifier}

We have shown in Section~\ref{sec:problem-formulation} that the fundamental task of NPN classification can be transformed into the task of predicting NPN actions that improve the Boolean function lexicographically.
In this section, we present our proposed GNN-NPN action predictor, which is the core of our workflow in Figure~\ref{fig:gnn-npn-workflow}.

\subsection{State Representation}
\label{sec:repr}

We adopt AIGs as the input representation for GNN-NPN.
This choice is motivated by two considerations.
First, AIGs are widely used in logic synthesis workflows and can be extracted directly from practical benchmarks.
Second, unlike canonical representations such as truth tables, AIGs preserve structural regularities such as symmetry, which are exploited by existing NPN classification methods~\cite{zhou2019fast} and can also be exploited by graph learning.

Each node in the AIG is assigned a discrete type label indicating whether it is a PI, \texttt{AND} node, or \texttt{NOT} node.
This type label is used to select the appropriate propagation and decoding modules in the heterogeneous GNN described below.
The hidden state \(h_i\in\mathbb{R}^d\) of every node \(i\) is initialized to the zero vector.

\subsection{Model Architecture}

We propose a bi-directional heterogeneous DAG-structured GNN model architecture for our GNN-NPN action predictor, which is illustrated in Fig.~\ref{fig:gnn-npn-workflow}.
Compared to previous works that use DAG-structured GNNs to model AIGs~\cite{shi2023deepgate2}, our proposed architecture adds backward propagation and thereby improves the model's ability to predict PI-related tasks such as input negation and input permutation.
The architecture is also heterogeneous, in the sense that it uses a separate parameter set for each combination of propagation direction and destination-node type.
This model can be applied to AIGs with arbitrary numbers of inputs and gates.

The proposed GNN-NPN architecture processes an AIG in three stages.
First, it initializes a \(d\)-dimensional hidden state \(h_i\) for every node \(i\) to zero.
Second, it performs \(R\) rounds of message passing, where each round consists of a forward sweep that follows the topological order of the AIG and a backward sweep that follows the reverse topological order.
Finally, three multi-layer perceptron (MLP) decoder heads are applied to the final node embeddings:
\begin{enumerate}
    \item an input-negation decoder that outputs \(l^{\mathrm{PI\text{-}Neg}}_j\) for each PI \(j\),
    \item an output-negation decoder that outputs \(l^{\mathrm{PO\text{-}Neg}}\) for the PO,
    \item a permutation decoder that outputs \(l^{\mathrm{Perm}}_j\) for each PI \(j\).
\end{enumerate}
Together, these three heads define the NPN action \(a=(p,\pi,q)\).

The propagation uses five heterogeneous relation types:
\begin{enumerate}
    \item forward updates to \texttt{AND} nodes,
    \item forward updates to \texttt{NOT} nodes,
    \item backward updates to \texttt{AND} nodes,
    \item backward updates to \texttt{NOT} nodes,
    \item backward updates to PI nodes.
\end{enumerate}
Each relation type uses the same message-passing rule but with an independent set of parameters.
Let \(t\) denote a propagation type, and let \(\mathcal{N}^{(t)}(i)\) be the neighbors of node \(i\) participating in type-\(t\) aggregation at the current propagation level.
The update is written schematically as
\begin{equation}
    {w^{(t)}_{ij}}=\mathrm{Softmax}\left\{\left.\frac{(W_q^{(t)}h_i)^\top(W_k^{(t)}h_j)}{\sqrt{d}}\right|j\in\mathcal{N}^{(t)}(i)\right\},
\end{equation}
\begin{equation}
    h_i\leftarrow \mathrm{GRU}^{(t)}\left(h_i,\sum_{j\in\mathcal{N}^{(t)}(i)}{w^{(t)}_{ij}}W_v^{(t)}h_j\right),
\end{equation}
where \(W_q^{(t)}\), \(W_k^{(t)}\), \(W_v^{(t)}\), and \(\mathrm{GRU}^{(t)}\) are the type-specific query, key, value, and update parameters, respectively.

\subsection{Action Sampling}

From the decoder outputs, we sample NPN transformations \(a=(p,\pi,q)\) with a temperature \(\tau\) to control the randomness.
Larger \(\tau\) encourages exploration while smaller \(\tau\) leads to more deterministic sampling, and \(\tau=0\) results in greedy deterministic sampling.
We use \(\tau=1.0\) during training to encourage exploration and use deterministic sampling (\(\tau=0\)) for the single-trial evaluation settings.

To sample input negation, we transform the PI-negation logits to probabilities using the sigmoid function \(\sigma\) and sample independently from Bernoulli distributions:
\begin{equation}
    p_j\sim\mathrm{Bernoulli}\left(\sigma(l^{\mathrm{PI\text{-}Neg}}_j/\tau)\right),\quad j=0,1,\dots,n-1.
\end{equation}
Similarly, the output negation bit is sampled as
\begin{equation}
    q\sim\mathrm{Bernoulli}\left(\sigma(l^{\mathrm{PO\text{-}Neg}}/\tau)\right).
\end{equation}

To sample the permutation, we use the Plackett-Luce model~\cite{plackett1975analysis} from the most significant bit (MSB) to the least significant bit (LSB).
The permutation decoder outputs one score \(l^{\mathrm{Perm}}_j\) per PI, and the same score vector is reused at every step.
At step \(k\), the scores are re-normalized over the currently available PI set \(\mathrm{PI}_k^{\mathrm{Avail}}\), i.e., the PIs that have not been selected in earlier steps.
\begin{equation}
    \pi_{k}\sim\mathrm{Categorical}\left(\mathrm{Softmax}\left(\left.l^{\mathrm{Perm}}_{j}/\tau\right|j\in \mathrm{PI}_k^{\mathrm{Avail}}\right)\right).
\end{equation}
After sampling \(\pi_k\), the selected PI is removed from \(\mathrm{PI}_k^{\mathrm{Avail}}\).

\subsection{Training Algorithms}
\label{sec:training}

The goal of action prediction is to optimize the Boolean function towards its canonical representation, that is, the NPN-equivalent Boolean function with the minimal lexicographical truth table value.

We consider two training strategies.
The first is a supervised baseline derived from exact NPN actions.
Input negation and output negation are treated as binary classification tasks, while permutation learning is decomposed into multiple binary classification tasks over all PI pairs, where each label indicates the relative order of the two PIs in the exact permutation.

The second, and primary, strategy is reinforcement learning based on the quality of the sampled action. We design the reward to reflect the reduction in lexicographical value from the original function \(f\) to the transformed function \(f'=a(f)\).
Specifically, we normalize the lexicographic improvement to \(\left(-1, 1\right)\):
\begin{equation}
    \label{eq:reward}
    r\left(f'|f\right)=\frac{\mathrm{lex}(f)-\mathrm{lex}(f')}{2^{2^{|\mathrm{PI}|}}}.
\end{equation}

We use Group Relative Policy Optimization (GRPO)~\cite{shao2024deepseekmath} to optimize the policy.
For each Boolean function, we sample a group of \(G\) NPN actions and compare the rewards \(\mathbf{r}=(r_1,r_2,\dots,r_G)\) within the group to estimate the advantage of each action.

\begin{equation}
        \hat{A}_{i}=\frac{r_i-\mathrm{mean}(\mathbf{r})}{\mathrm{std}(\mathbf{r})},\quad i=1,2,\dots,G.
\end{equation}

The weights of the GNN-NPN action predictor \(\theta\) are updated using the Proximal Policy Optimization~\cite{schulman2017proximal} objective for \(K\) epochs per group, with a clipping factor \(\epsilon\) to avoid drastic updates and an entropy term \(H\)~\cite{williams1991function} to regularize the policy.
The loss function is

\begin{equation}
\begin{aligned}
    L_{\theta} = -&\frac{1}{B}\sum_{i=1}^B\left[\min\left( \lambda_i \hat{A}_{i},\text{clip}\bigl(\lambda_i,\, 1-\epsilon,\, 1+\epsilon\bigr)\hat{A}_i\right)\right]\\
    -&\beta\cdot\frac{1}{B}\sum_{i=1}^B H_{\theta}(a_i),
\end{aligned}
\end{equation}
where \(\lambda_i=P_{\theta}(a_i)/P_{\theta_0}(a_i)\) is the importance sampling ratio between the action probabilities after and before the update.

\section{Graph Rewriting for NPN Classification}
\label{sec:graph-representation}

Although AIG is an expressive intermediate representation for the GNN in Section~\ref{sec:gnn-classifier}, it is not a canonical form and inevitably introduces structural bias that can hamper the model's performance.
The same Boolean function may be realized by many different graphs.
For instance, the function {$f_2$} in Fig.~\ref{fig:gnn-npn-workflow} can be implemented using either {$G_2$} or {$G_3$}, each with a different graph structure.
This structural variability poses a challenge for GNNs to learn consistent representations for NPN-equivalent functions, as the same function can be represented by different graph topologies.

In this section, we develop two graph rewriting algorithms designed to mitigate this structural variability.
In an ideal scenario, logic network rewriting would be both NPN-invariant and exhibit perfect reachability: it would normalize all equivalent functions into a single structural form, differing only in input ordering and inversions.
However, such an idealized oracle does not exist, which motivates the use of a learned predictor as introduced in Section~\ref{sec:gnn-classifier}.
Nevertheless, even without achieving perfectly identical graphs, applying specific rewriting rules can reduce structural differences among graphs within the same NPN class.
By increasing graph similarity, the rewriting step can improve NPN action prediction and downstream NPN classification accuracy.

\subsection{Symmetry Recovery}\label{subsec:symmetry-recovering}
After graph initialization, we introduce a mapping procedure in Algorithm~\ref{alg:ambiguity-mapping} that scans the graph and recovers symmetries broken by structural bias.
The key idea is to identify associative chains of \texttt{AND} gates with uniform polarity and merge them into higher-order super nodes.
This transformation preserves functionality while standardizing graph structure, making it easier for GNNs to learn symmetry-aware representations as discussed in Section~\ref{sec:repr}.
The merged node is treated as an \texttt{AND}-type node by the GNN and therefore reuses the same message-passing parameters as \texttt{AND} nodes.

\begin{figure}[tb]
    \centering
    \begin{subfigure}[b]{0.3\linewidth}
        \centering
        \begin{tikzpicture}[scale=0.48, every node/.style={transform shape}]
       \node (a) at (0,0) {\large $x_0$};
       \node (b) at (1,0) {\large $x_1$};
       \node (c) at (2,0) {\large $x_2$};

            \node[draw, circle, minimum width=0.6cm, minimum height=0.4cm] (and1) at (0.5,1) {$\wedge$};
            \node[draw, circle, minimum width=0.6cm, minimum height=0.4cm] (and2) at (1.25,2) {$\wedge$};

            \node (f) at (1.25,3) {\large $f$};

            \draw (a) -- (and1);
            \draw (b) -- (and1);
            \draw (and1) -- (and2);
            \draw (c) -- (and2);
            \draw (and2) -- (f);
        \end{tikzpicture}
        \caption{ $(x_0 \wedge x_1)\wedge x_2$}
    \end{subfigure}
    \hfill
    \begin{subfigure}[b]{0.3\linewidth}
        \centering
        \begin{tikzpicture}[scale=0.48, every node/.style={transform shape}]
       \node (a) at (0,0) {\large $x_0$};
       \node (b) at (1,0) {\large $x_1$};
       \node (c) at (2,0) {\large $x_2$};

            \node[draw, circle, minimum width=0.6cm, minimum height=0.4cm] (and1) at (1.5,1) {$\wedge$};
            \node[draw, circle, minimum width=0.6cm, minimum height=0.4cm] (and2) at (0.75,2) {$\wedge$};

            \node (f) at (0.75,3) {\large $f$};

            \draw (b) -- (and1);
            \draw (c) -- (and1);
            \draw (and1) -- (and2);
            \draw (a) -- (and2);
            \draw (and2) -- (f);
        \end{tikzpicture}
        \caption{ $x_0 \wedge (x_1 \wedge x_2)$}
    \end{subfigure}
    \hfill
    \begin{subfigure}[b]{0.3\linewidth}
        \centering
        \begin{tikzpicture}[scale=0.48, every node/.style={transform shape}]
       \node (a) at (0,0) {\large $x_0$};
       \node (b) at (1,0) {\large $x_1$};
       \node (c) at (2,0) {\large $x_2$};

            \node[draw, circle, minimum width=0.8cm, minimum height=0.5cm] (and3) at (1,1.5) {$\wedge$};

            \node (f) at (1,2.7) {\large $f$};

            \draw (a) -- (and3);
            \draw (b) -- (and3);
            \draw (c) -- (and3);
            \draw (and3) -- (f);
        \end{tikzpicture}
        \caption{ $\bigwedge (x_0,x_1,x_2)$}
    \end{subfigure}

    \caption{Structural ambiguity (a, b) and resolution (c) via node merging.}
    \label{fig:ambiguity-removal}
\end{figure}
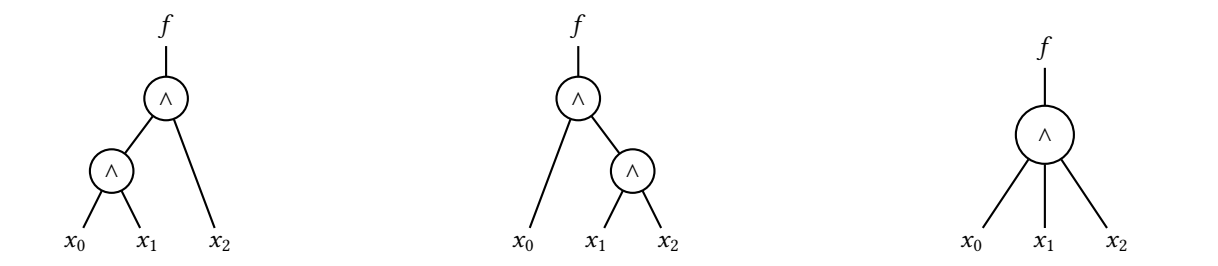

Figure~\ref{fig:ambiguity-removal} shows an example where the same three-input AND function can be represented either as $(x_0 \wedge x_1) \wedge x_2$ or as $x_0 \wedge (x_1 \wedge x_2)$.
Although these representations are functionally equivalent, their different tree structures produce different graph topologies.
In our framework, we address this issue by unifying these structures and grouping the entire associative chain into a single super node.

\subsection{Ambiguity-Oriented Graph Rewriting}\label{subsec:ambiguity-rewriting}

\begin{algorithm}[tb]
\caption{Symmetry-Recovery Mapping}
\label{alg:ambiguity-mapping}
\footnotesize
\begin{algorithmic}[1]
\REQUIRE Logic network $G$
\REQUIRE Predicate $\textsc{Mergeable}(u,v)$ for associative \texttt{AND}-type merging
\FORALL{node $n$ in $G$}
    \IF{$\text{type}(n)=\texttt{AND}$ and $n$ is not yet merged}
        \STATE $S \gets \{n\}$
        \STATE $\mathcal{Q} \gets [n]$
        \WHILE{$\mathcal{Q}$ is not empty}
            \STATE Pop a node $u$ from $\mathcal{Q}$
            \FORALL{fanin $l\in \delta^-(u)$ of $u$}
                \IF{$l\notin S$ and $\textsc{Mergeable}(u,l)$}
                    \STATE $S \gets S \cup \{l\}$
                    \STATE Push $l$ into $\mathcal{Q}$
                \ENDIF
            \ENDFOR
        \ENDWHILE
        \IF{$|S| > 1$}
            \STATE Replace nodes in $S$ with one \texttt{AND}-type super node
        \ENDIF
    \ENDIF
\ENDFOR
\end{algorithmic}
\end{algorithm}

\begin{algorithm}[tb]
\caption{Ambiguity-Oriented Rewriting}
\label{alg:ambiguity-rewriting}
\footnotesize
\begin{algorithmic}[1]
\REQUIRE Logic network $G$
\REQUIRE Boolean optimizer that returns equivalent candidates for a given window
\STATE Initialize: $G' \gets G$
\FORALL{window $W \subseteq G'$}
    \STATE Let $C_0$ be the current subgraph in $W$
    \STATE $\mathcal{C} \gets$ functionally equivalent candidates($C_0$)
    \STATE Initialize: $\text{best\_cost} \gets \text{Cost}(C_0)$, $\text{best\_cand} \gets C_0$
    \FORALL{candidate circuit $C_i \in \mathcal{C}$}
        \STATE $s_i \gets \text{Cost}(C_i)$
        \IF{$s_i < \text{best\_cost}$}
            \STATE $\text{best\_cost} \gets s_i$
            \STATE $\text{best\_cand} \gets C_i$
        \ENDIF
    \ENDFOR
    \STATE Replace $C_0$ in $G'$ with $\text{best\_cand}$
\ENDFOR
\RETURN $G'$
\end{algorithmic}
\end{algorithm}

While symmetry recovery focuses on collapsing structurally redundant patterns, our ambiguity-oriented graph rewriting actively reshapes the logic network.
This approach both shortens the propagation path of the GNN and increases opportunities for effective symmetry recovery.

More specifically, our algorithm identifies local rewriting opportunities using a window-based peephole approach. For each window selected for optimization, we employ the resynthesis engine to generate a set of \emph{candidate circuits} that preserve functionality. Our task is then to evaluate each candidate and select the one that minimizes structural ambiguity and aligns best with the surrounding context.

To guide this decision, we define a cost metric for each candidate circuit based on its potential to merge with neighboring gates into a larger super gate. Formally, for each candidate circuit \( C_i \), we define the ambiguity cost as:
\[
\text{Cost}(C_i) = \max_{n \in C_i}\, \text{level}(n),
\]
where \(\text{level}(n)\) represents the minimum number of propagation levels in the GNN from node \(n\) to any primary input from which NPN actions are predicted. More precisely,
\[
\text{level}(n) =
\begin{cases}
0, & \text{if } n\in\mathrm{PI}, \\[4pt]
\displaystyle \min_{l\in \delta^-(n)}
\begin{cases}
\text{level}(l), & \text{if } \text{match}(n, l), \\[4pt]
\text{level}(l)+1, & \text{otherwise,}
\end{cases}
& \text{if } n\notin\mathrm{PI}.
\end{cases}
\]
Here, \(\text{match}(n, l)\) is true if node \(n\) and its fanin \(l\) can be merged by the symmetry-recovery rule, i.e., they have compatible \texttt{AND}-type semantics and the connecting edge satisfies the required polarity.

Overall, the rewriting flow enhances the effectiveness of symmetry recovery, reduces the number of layers between nodes and primary inputs, and improves the efficiency of message passing.

\section{Experimental Results}
\label{sec:exp}

In this section, we evaluate the proposed GNN-NPN workflow from three perspectives: action prediction, canonicalization quality, and practical NPN classification performance.
We organize the experiments around the following three research questions (RQs).

\begin{enumerate}
    \item [RQ1]: How well can GNN-NPN predict NPN actions that improve a Boolean function toward its canonical representative?
    \item [RQ2]: How much can the predicted actions improve NPN canonicalization quality compared with existing heuristic methods?
    \item [RQ3]: How does the full GNN-NPN workflow perform on practical logic synthesis benchmarks in terms of NPN classification quality and scalability?
\end{enumerate}

\subsection{Implementation Details}

During training, we generate random AIGs corresponding to 6-input Boolean functions on the fly.
For the small-scale evaluations in RQ1 and RQ2, we use a separate set of 1000 random 6-input Boolean functions, for which the exact canonical representative and exact NPN action are computed using the \texttt{exact\_npn} implementation in the kitty library~\cite{soeken2018epfl}.
For the practical benchmark evaluation in RQ3, we extract subcircuits with 6--16 inputs from the 20 circuits in the EPFL Combinational Benchmark Suite~\cite{amaru2015epfl}, covering both arithmetic and control circuits.
The subcircuits are generated using reconvergence-driven cuts implemented in mockturtle~\cite{soeken2018epfl}.

Our baselines include two heuristics in the kitty library~\cite{soeken2018epfl} and eleven NPN classification algorithms in ABC~\cite{brayton2010abc}.
The two kitty heuristics are the flip-swap and sifting heuristics proposed {in}~\cite{huang2013fast}. {Both greedily apply single-bit negations and two-bit swaps until no further lexicographical improvement is found.}
{The \texttt{testnpn} command in ABC provides exact and heuristic NPN classification algorithms selected through its \texttt{-A} option~\cite{huang2013fast,petkovska2016fast,zhou2019fast}. They are non-learning methods that operate on truth tables rather than AIGs.}

For all experiments, we set the embedding dimension \(d=128\) and the number of propagation rounds \(R=3\).
We use three-layer multi-layer perceptrons (MLPs) for decoding logits.
For the GRPO algorithm, the clipping parameter is \(\epsilon=0.2\), the entropy loss weight is \(\beta=10^{-4}\), the number of PPO epochs is \(K=4\), the mini-batch size is \(B=64\) and the group size is \(G=16\). Adam optimizer~\cite{kingma2014adam} with {a} learning rate \(\alpha=2\times10^{-5}\) and cosine learning rate annealing~\cite{loshchilov2016sgdr} is used to train our models for 10K steps.
All of our experiments are performed on server machines with AMD EPYC 7V13 64-Core Processors and AMD Instinct MI210 GPUs. The models are built using PyTorch~\cite{pytorch} and PyTorch {Geometric}~\cite{pyg}.

\subsection{Evaluation of Action Prediction}
\label{subsec:exp-repr}
\begin{table*}[tb]
    \centering
    \caption{{Evaluation of NPN action prediction for different training algorithms and GNN architecture variants.} The columns from left to right are: the accuracy of prediction for input negation \(p\), output negation \(q\), input permutation \(\pi\) (pairwise and exact), reward {(defined in Equation~\ref{eq:reward})} achieved by the predicted action, and the ratio of canonical representative \(\hat{f}\) achieved directly and with flip-swap and sifting post-processing, respectively. {The architecture ablations use GRPO with graph rewriting.}}
    \label{tab:action}
    \begin{tabular}{lrrrrrrrr}
        \toprule
            &\(p\)\% & \(q\)\%& \(\pi\)\% (Pairwise) & \(\pi\)\% (Exact) & \(r\) & \(\hat{f}_{\mathrm{direct}}\)\%  &\(\hat{f}_{\mathrm{Flip-swap}}\)\%& \(\hat{f}_{\mathrm{Sifting}}\)\% \\
         \midrule
        \multicolumn{9}{l}{{\textbf{Training Algorithm}}}\\
        \midrule
      Supervised&87.7\%& 96.9\%& 71.07\%& 4.20\%&0.2232 & 22.30\% &60.78\%&74.66\%\\
   GRPO&77.69\%& 95.62\%& 72.12\%& 5.70\%&0.2290 &19.59\% &60.78\%&76.49\%\\
   GRPO + Rewriting&77.47\%& 95.83\%& 72.31\%& 5.71\%&0.2773 & 20.95\% &60.78\%&79.70\%\\
        \midrule
        \multicolumn{9}{l}{{\textbf{GNN Architecture Ablations}}}\\
        \midrule
        {Uni-directional}&{47.63\%}&{68.28\%}&{50.39\%}&{0.12\%}&{0.1962}&{0.12\%}&{47.50\%}&{63.78\%}\\
        {Homogeneous}&{66.04\%}&{81.07\%}&{65.96\%}&{2.16\%}&{0.2361}&{0.67\%}&{58.24\%}&{67.10\%}\\
        \bottomrule
    \end{tabular}
\end{table*}

To answer RQ1, we evaluate action prediction on the 1000 random 6-input Boolean functions, where exact NPN actions are available.

Table~\ref{tab:action} compares supervised training, GRPO training, and GRPO training with graph rewriting.
All three models achieve strong PI/PO negation accuracy.
For permutation prediction, the exact permutation accuracy is low, but the pairwise permutation accuracy exceeds 70\% for all methods, indicating that the predicted ordering is often locally consistent even when the full permutation is not exactly recovered.
This behavior is important because our downstream flow can refine a predicted action with local search.
It is also worth noting that the ratio of canonical representative \(\hat{f}_{\mathrm{direct}}\)\% directly achieved with the predicted action is higher than the exact permutation accuracy \(\pi\)\% (Exact), which indicates that some symmetry exists among PIs.

Comparing supervised learning and GRPO, supervised training yields slightly higher direct action-prediction accuracy and initial canonicalization success, whereas GRPO achieves better post-processed canonicalization.
This trend is consistent with the idea that reward-based training better aligns the predictor with the downstream objective of improving the representative, especially for permutation prediction.
Adding rewriting further improves reward and canonicalization-related metrics, which supports using rewriting in the default workflow adopted later. In the following experiments, we will {use} the GRPO algorithm with rewriting as the default training algorithm for GNN-NPN, as it achieves the best performance in action prediction and canonicalization.

{To isolate the benefits of the proposed architecture, we compare it with uni-directional and homogeneous variants trained using the same GRPO and graph rewriting setup. Removing backward propagation yields near-random PI-related predictions, with input-negation and pairwise permutation accuracies {dropped to} 47.63\% and 50.39\%, respectively. The homogeneous variant also degrades all canonicalization metrics, demonstrating the benefits of both bidirectional propagation and type-specific parameters.}

\subsection{Evaluation of NPN Canonicalization}
\label{subsec:exp-classification}
\begin{figure*}[tb]
    \centering
    \begin{subfigure}[b]{0.45\textwidth}
        \centering
        \includegraphics[width=0.7\textwidth]{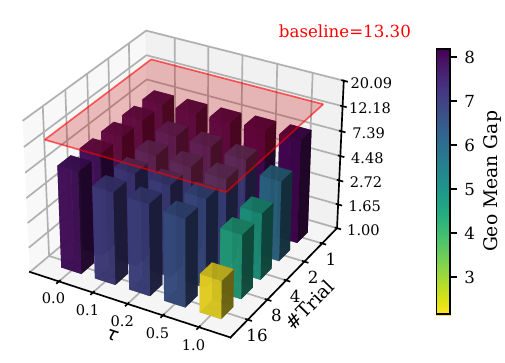}
        \caption{Flip-swap post-processing.}
    \end{subfigure}
    \hfill
    \begin{subfigure}[b]{0.45\textwidth}
        \centering
        \includegraphics[width=0.7\textwidth]{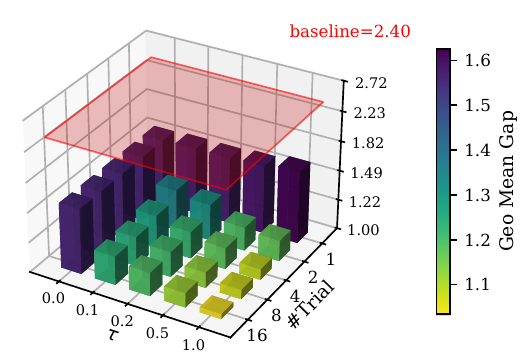}
        \caption{Sifting post{-}processing.}
    \end{subfigure}
    \caption{Comparison of canonicalization gaps using different post-processing, temperature \(\tau\), and number of trials.}
    \label{fig:canon}
\end{figure*}

To answer RQ2, we again evaluate on the same 1000 random 6-input Boolean functions used in RQ1.
The ground-truth canonical representative \(\hat{f}\) is obtained by exact NPN canonicalization using kitty's \texttt{exact\_npn} implementation~\cite{soeken2018epfl}.
We measure the quality of the representative \(f'\) derived by our workflow using
\begin{equation}
    \mathrm{gap}(f',\hat{f})=\frac{\mathrm{lex}(f')}{\mathrm{lex}(\hat{f})},
\end{equation}
where smaller values are better.

Unless otherwise stated, the reported GNN-NPN results in this subsection use the model trained with GRPO and graph rewriting, and each sampled action is followed by flip-swap or sifting as post-processing for local refinement.

To further enhance the quality of the derived representative, we sample multiple NPN actions from the logits produced by the GNN-NPN predictor.
This is inexpensive because it samples from the predicted distribution without additional model inference.
We denote the number of sampled actions \(\#Trials\).
When \(\# Trials=1\), the sampling is deterministic and corresponds to greedy decoding \(\tau=0\).
When \(\# Trials>1\), the sampling is stochastic with a temperature \(\tau>0\) that controls the randomness of sampling. Larger \(\tau\) encourages more exploration of the action space.

In Figure~\ref{fig:canon}, we perform grid searches on the number of trials and temperature.
Compared to the two baselines that directly apply the flip-swap and sifting post-processing to {the} original {function}, GNN-NPN consistently achieves lower gap values, even with a single trial.
With more trials and a suitable temperature, GNN-NPN further improves the derived representative and reaches substantially smaller gaps.
These results show that the predictor logits provide a useful ranking over candidate NPN actions and that multi-trial sampling is an effective way to exploit that information.

\subsection{Evaluation of NPN Classification}
\begin{table*}[tb]
    \centering
    \caption{Performance of NPN classification, quantified by number of NPN equivalent classes reduced from subcircuits of the EPFL Combinatorial Benchmark Suite~\cite{amaru2015epfl}, the smaller the better. The timeout is 10 hours. The best results are in bold.}
    \label{tab:epfl}

    \begin{tabular}{llrrrrrr}
        \toprule
        Method & Note & 6-bit & 8-bit & 10-bit & 12-bit & 14-bit & 16-bit \\

        \midrule

        None & Number of functions & 403672 & 370269 & 360215 & 354395 & 339435 & 334501 \\

        \midrule
        
        \multirow{2}{*}{kitty~\cite{soeken2018epfl}} & Flip-swap~\cite{huang2013fast} & 2028 & 10507 & 20950 & 31717 & 40537 & 46774 \\
        & Sifting~\cite{huang2013fast} & 1828 & 10301 & 20573 & 31088 & 39578 & 45344 \\

        \midrule
        
        \multirow{11}{*}{ABC \texttt{testnpn}~\cite{brayton2010abc}} & \texttt{-A 0} (uniquify) & 12847 & 30099 & 39942 & 55349 & 63054 & 66993 \\
        & \texttt{-A 1} (exact) & \textbf{1253} & \textbf{7829} & timeout & timeout & timeout & timeout \\
        & \texttt{-A 2} & 2151& 10640& 21849& 36470& 45594&50070\\
        & \texttt{-A 3} & 1321& 8001& 17794& 27442& 35208&40172\\
        & \texttt{-A 4} & 1291& 7935& 17644& 27116& 34742&39688\\
        & \texttt{-A 5} & 1314& 8019& 17869& 27938& 35827&40904\\
        & \texttt{-A 6} & 4503& 17189& 27376& 39591& 48232&54075\\
        & \texttt{-A 7} & 1314& 8019& 17869& 27938& 35827&40904\\
        & \texttt{-A 8} & \textbf{1253}& 7834& 17490& 26981& 34552&39419\\
        & \texttt{-A 9} & \textbf{1253}& 7834& 17490& 26981& 34552&39419\\
        & \texttt{-A 10} (exact) & \textbf{1253}& \textbf{7829}& \textbf{17430}& \textbf{26495}& timeout&timeout\\

        \midrule
        
        GNN-NPN (Ours) & \#Trial=16, sifting post-processed& \textbf{1253}& \textbf{7829}& 17474& 26512& \textbf{33856}&\textbf{38939}\\
        \bottomrule
    \end{tabular}
\end{table*}

\begin{table*}[!t]
    \centering
    \begin{minipage}[t]{0.56\textwidth}
    \centering
    \caption{Runtime breakdown and memory footprint of GNN-NPN on EPFL subcircuits at different bit widths.}
    \label{tab:breakdown}
        \begin{tabular}{lrrrrrr}
            \toprule
            Bit Width & 6-bit & 8-bit & 10-bit & 12-bit & 14-bit & 16-bit \\
            \midrule
            \multicolumn{7}{l}{Runtime (s)} \\
            \midrule
            Graph Rewriting & 3.5 & 22.1 & 38.3 & 79.8 & 150.5 & 294.3 \\
            Model Inference & 93.75 & 319.01 & 611.15 & 1245.75 & 1576.11 & 2129.56 \\
            Action Sampling & 0.03 & 0.06 & 0.12 & 0.26 & 0.40 & 0.48 \\
            Post-Processing & 0.30 & 0.63 & 1.50 & 4.62 & 19.81 & 95.97 \\
            \midrule
            \multicolumn{7}{l}{Memory Footprint (MB)} \\
            \midrule
            CPU Memory & 3.8 & 32.1 & 105.8 & 511.5 & 2227.4 & 9338.3 \\
            GPU Memory & 117.9 & 177.7 & 246.6 & 378.5 & 495.5 & 618.8 \\
            \bottomrule
        \end{tabular}%
\end{minipage}
    \hfill
    \begin{minipage}[t]{0.42\textwidth}
    \vspace{0pt}
    \centering
    \includegraphics[width=0.7\linewidth]{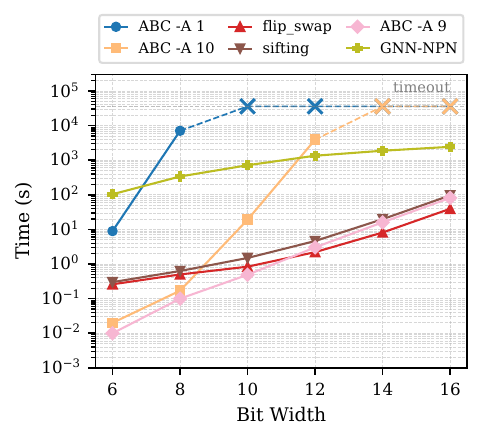}
    \makeatletter\def\@captype{figure}\makeatother
    \caption{Scalability of exact, heuristic, and GNN-NPN methods on EPFL subcircuits with 6--16 input bits.}
    \label{fig:scalability}
\end{minipage}
\end{table*}

Finally, to answer RQ3, we evaluate how well GNN-NPN groups practical Boolean functions into NPN-equivalence classes.
This is of practical value in modern logic synthesis workflows, where a precomputed library is indexed by a canonical representative for each function class.
At runtime, each lookup tests whether the current function matches one of these representatives, and, on a match, replaces the existing network fragment with the optimized, precomputed structure.
Therefore, better NPN classification quality improves optimization opportunities and reduces library size.

We extract subcircuits of 6--16 inputs from the 20 circuits in the EPFL Combinational Benchmark Suite~\cite{amaru2015epfl} and compare the GNN-NPN workflow against all baselines.
The metric is the number of distinct NPN-equivalence classes obtained after canonicalization, smaller values are better.
This metric is appropriate here because all compared methods map each function only to representatives within its own NPN class; therefore, errors manifest as failing to merge equivalent functions, which increases the final number of classes.
We set a 10{-}hour timeout.
As shown in Table~\ref{tab:epfl}, GNN-NPN with 16 trials and sifting as post-processing achieves performance close to the exact ABC methods (\texttt{-A 1} and \texttt{-A 10}) on problem sizes where exact methods remain tractable, while continuing to scale to 16-input functions.
The exact methods stop at 8 inputs and 12 inputs, respectively.
At the same time, GNN-NPN produces fewer classes than all heuristic baselines on larger problem sizes.
These results support the central claim of the paper: the proposed workflow closes much of the quality gap to exact methods while retaining the scalability advantages of heuristic methods.

We further analyze the scalability of GNN-NPN by plotting the time-bit curve in Figure~\ref{fig:scalability}.
The exact methods (\texttt{-A 1} and \texttt{-A 10}) scale super-exponentially and quickly become intractable as the number of inputs increases.
Heuristic methods such as flip-swap, sifting, and \texttt{-A 9} (the best{-}performing heuristic in ABC) scale more favorably, but their runtime still increases substantially with the input size.
In contrast, GNN-NPN shows a much more favorable scalability, with runtime increasing only slightly as the number of bits increases.
This trend is consistent with the graph-based design of GNN-NPN, which avoids exhaustive truth-table manipulation and instead operates on compact logic-network representations.

{Table~\ref{tab:breakdown} further reports the runtime breakdown and memory footprint of GNN-NPN. The end-to-end runtime includes graph rewriting, model inference with a batch size of 256 on GPU, action sampling, and post-processing. Model inference is the dominant stage at every bit width.}

\section{Conclusion and Discussion}
In this paper, we introduced a learning-based formulation for NPN classification by casting it as NPN action prediction rather than directly predicting class labels or canonical representatives.
Based on this formulation, we proposed a graph-based workflow that represents Boolean functions as AIGs, predicts NPN actions with a bidirectional heterogeneous GNN, and reduces structural ambiguity through graph rewriting.
Experimental results showed that the proposed predictor learns useful NPN actions, improves canonicalization quality over heuristic baselines, and achieves practical classification quality close to exact methods on tractable problem sizes while scaling to larger Boolean functions.

A natural question is how to take advantage of the better NPN matching capability offered by GNN-NPN. 
In our initial experiment, we performed cut enumeration in a given logic network using cut sizes ranging from 6 to 16 following the method in~\cite{10.1145/296399.296425}, performed NPN-matching against a highly optimized library of small size logic functions~\cite{iwls_contest}, and replaced whenever we see an improvement.  However, the gain from this simple approach is limited so far.  The results from sifting~\cite{huang2013fast} are similar, despite the fact that GNN-NPN can reduce the number of NPN classes by 23\%.   More research is needed to better utilize the capability offered by GNN-NPN. 
We also would like to point out that generalization of GNN-NPN beyond EPFL-derived subcircuits has not been verified and may degrade on out-of-distribution circuits.

\begin{acks}
This work was partially supported by NSF grants 2410716 and 2533041, and the
CDSC industrial partners (\url{https://cdsc.ucla.edu/partners/}).
\end{acks}

\clearpage
\bibliographystyle{ACM-Reference-Format}
\bibliography{gnn-npn}

\end{document}